# International Standard for a Linguistic Annotation Framework

**Nancy Ide**
Dept. of Computer Science
Vassar College
Poughkeepsie, New York 12604-0520 USA
`ide@cs.vassar.edu`

**Laurent Romary**
Equipe Langue et Dialogue
LORIA/INRIA
Vandoeuvre-lès-Nancy
FRANCE
`romary@loria.fr`

## Abstract

This paper describes the Linguistic Annotation Framework under development within ISO TC37 SC4 WG1. The Linguistic Annotation Framework is intended to serve as a basis for harmonizing existing language resources as well as developing new ones.

## 1 Introduction

Language resources are bodies of electronic language data used to support research and applications in the area of natural language processing. Typically, such data are enhanced (annotated) with linguistic information such as morpho-syntactic categories, syntactic or discourse structure, co-reference information, etc.; or two or more bodies may be aligned for correspondences (e.g., parallel translations, speech signal and transcription).

Over the past 15-20 years, increasingly large bodies of language resources have been created and annotated by the language engineering community. Certain fundamental representation principles have been widely adopted, such as the use of stand-off annotation (Ide and Priest-Dorman, 1996), use of XML, etc., and several attempts to provide generalized annotation mechanisms and formats have been developed (e.g., XCES (Ide, *et al.,* 2000), annotation graphs (Bird and Liberman, 2001)). However, it remains the case that annotation formats often vary considerably from resource to resource, often to satisfy constraints imposed by particular processing software. The language processing community has recognized that commonality and interoperability are increasingly imperative to enable sharing, merging, and comparison of language resources.

To provide an infra-structure and framework for language resource development and use, the International Organization for Standardization (ISO) has formed a sub-committee (SC4) under Technical Committee 37 (TC37, Terminology and Other Language Resources) devoted to Language Resource Management. The objective of ISO/TC 37/SC 4 is to prepare international standards and guidelines for effective language resource management in applications in the multilingual information society. To this end, the committee is developing principles and methods for creating, coding, processing and managing language resources, such as written corpora, lexical corpora, speech corpora, dictionary compiling and classification schemes. The focus of the work is on data modeling, markup, data exchange and the evaluation of language resources other than terminologies (which have already been treated in other sub-committees of ISO/TC 37). The worldwide use of ISO/TC 37/SC 4 standards should improve information management within industrial, technical and scientific environments, and increase efficiency in computer-supported language communication.

At present, language professionals and standardization experts are not sufficiently aware of the standardization efforts being undertaken by ISO/TC 37/SC 4. Promoting awareness of future activities and rising problems, therefore, is crucial for the success of the committee, and will be required to ensure widespread adoption of the standards it develops. An even more critical factor for the success of the committee's work is to involve, from the outset, as many and as broad a range of potential users of the standards as possible.

Within ISO/TC 37/SC 4, a working group (WG1) has been established to develop a Linguistic Annotation Framework (LAF) that can serve as a basis for harmonizing existing language resources as well as developing new ones. In order to ensure that the framework is developed based on the input and consensus of the research community, a group of experts[1] was convened on November 21-22, 2002, at Pont-à-Mousson, France, to lay out the overall structure of the framework. Based on the determinations of the experts at the workshop, the general outlines of the Linguistic Annotation Framework have been defined. In this paper, we describe the LAF design as it has been developed so far, and solicit the input of other members of the community to inform its further development.

## 2  Background and rationale

The standardization of principles and methods for the collection, processing and presentation of language resources requires a distinct type of activity. Basic standards must be produced with wide-ranging applications in view. In the area of language resources, these standards should provide various technical committees of ISO, IEC and other standardizing bodies with the groundwork for building more precise standards for language resource management.[2]

The need for harmonization of representation formats for different kinds of linguistic information is critical, as resources and information are more and more frequently merged, compared, or otherwise utilized in common systems. This is perhaps most obvious for processing multi-modal information, which must support the fusion of multimodal inputs and represent the combined and integrated contributions of different types of input (e.g., a spoken utterance combined with gesture and facial expression), and enable multimodal output (see, for example, Bunt and Romary, 2002). However, language processing applications of any kind require the integration of varieties of linguistic information, which, in today's environment, come from potentially diverse sources. We can therefore expect use and integration of, for example, syntactic, morphological, discourse, etc. information for multiple languages, as well as information structures like domain models and ontologies.

We are aware that standardization is a difficult business, and that many members of the targeted communities are skeptical about imposing any sort of standards at all. There are two major arguments against the idea of standardization for language resources. First, the diversity of theoretical approaches to, in particular, the annotation of various linguistic phenomena suggests that standardization is at least impractical, if not impossible. Second, it is feared that vast amounts of existing data and processing software, which may have taken years of effort and considerable funding to develop, will be rendered obsolete by the acceptance of new standards by the community. Recognizing the validity of both of these concerns, WG1 does not seek to establish a single, definitive annotation scheme or format. Rather, the goal is to provide a framework for linguistic annotation of language resources that can serve as a reference or pivot for different annotation schemes, and which will enable their merging and/or comparison. To this end, the work of WG1 includes the following:

- analysis of the full range of annotation types and existing schemes, to identify the fundamental structural principles and content categories;

- instantiation of an abstract format capable of capturing the structure and content of linguistic annotations, based on the analysis in (1);

- establishment of a mechanism for formal definition of a set of reference content categories which can be used "off the shelf" or serve as a point of departure for precise definition of new or modified categories.

---

[1] Participants: Nuria Bel (Universitat de Barcelona), David Durand (Brown University), Henry Thompson (University of Edinburgh), Koiti Hasida (AIST Tokyo), Eric De La Clergerie (INRIA), Lionel Clement (INRIA), Laurent Romary (LORIA), Nancy Ide (Vassar College), Kiyong Lee (Korea University), Keith Suderman (Vassar College), Aswani Kumar (LORIA), Chris Laprun (NIST), Thierry Declerck (DFKI), Jean Carletta (University of Edinburgh), Michael Strube (European Media Laboratory), Hamish Cunningham (University of Sheffield), Tomaz Erjavec (Institute Jozef Stefan), Hennie Brugman (Max-Planck-Institut für Psycholinguistik), Fabio Vitali (Universite di Bologna), Key-Sun Choi (Korterm), Jean-Michel Borde (Digital Visual), Eric Kow (LORIA).
[2] This is particularly true for the two domains of Multimedia (ISO/IEC JTC1/SC 29/WG 11) and Education (ISO IEC/JTC1/SC 36)

- provision of both a set of guidelines and principles for developing new annotation schemes and concrete mechanisms for their implementation, for those who wish to use them.

By situating all of the standards development squarely in the framework of XML and related standards such as RDF, OWL, etc., we hope to ensure not only that the standards developed by the committee provide for compatibility with established and widely accepted web-based technologies, but also that transduction from legacy formats into XML formats conformant to the new standards is feasible.

## 3 General requirements for a linguistic annotation framework

### 3.1 Usage scenarios

Natural language processing (NLP) applications can be applied to *create* annotations for linguistic data by analyzing text, speech, and data representing other modalities to determine specific linguistic attributes and associate them with the segments of that data to which they apply. NLP applications also *use* linguistic annotations to facilitate language understanding and generation. Development of a standard linguistic annotation framework must proceed by considering both of these "views" on linguistic annotation, and integrating the two to ensure maximal inter-operability.

Annotation of linguistic data may involve multiple annotation steps, for example, morpho-syntactic tagging, syntactic analysis, entity and event recognition, semantic annotation, co-reference resolution, discourse structure analysis, etc. Annotation at higher linguistic levels typically relies on annotations at lower levels—that is, information at lower linguistic levels serves as input in the determination of higher-level annotation categories, so that annotation can be viewed as an *incremental* process. Depending on the application intended to use the annotations, lower-level annotations may or may not be preserved in a persistent format. That is, the output of the annotation software may consist solely of higher-level annotations, even though lower-level analysis has been performed. Note that many application programs—e.g., information extraction software—perform the analysis required for annotation of various linguistic features and utilize it internally to deliver the desired result, without preserving the annotation information.

The need to support annotations in the context of the Semantic Web is one of the most important considerations for development of the Linguistic Annotation Framework. Annotated corpora are, at present, primarily static entities used mainly for training annotation software, as well as for corpus linguistics and lexicography (which rely on annotated corpora to study language use). However, the advent of the Semantic Web and the development of supporting technologies will significantly alter the ways in which annotations are used and preserved in the future. In the context of the Semantic Web, annotations for a variety of (at least) higher-level linguistic and communicative features will be preserved in web-accessible form and used by software agents and other analytic software for inferencing and retrieval. This demands that the Linguistic Annotation Framework not only relies on web technologies (e.g., RDF, OWL) for representing annotations, but also that "layers' of annotations for the full range of annotation types (including named entities, time, space, and event annotation, annotation for gesture, facial expression, etc.) are at the same time *separable* (so that agents and other analytic software can access only those annotation types that are required for the purpose, and *mergeable* (so that two or more annotation types can be combined where necessary). They may also need to be *dynamic*, in the sense that new and/or modified information can be added as necessary.

Another increasingly important concern for LAF development is the handling of *streamed* data, wherein the processor analyzes input as it is encountered in a linear, time-bound sequence. Streamed data can be text, video, and audio, or might be a stream of sensor readings, satellite images, etc. This dictates that annotations to be attached to the data may be (temporarily) partial, especially where long-distance dependencies between seen and unseen segments of the data exist.

### 3.2 Requirements

To serve the goals of creation and use of linguistic annotation discussed above, we identify the fol-

lowing general requirements for a linguistic annotation framework:

*Expressive adequacy*. The framework must provide means to represent all varieties of linguistic information (and possibly also other types of information). This includes representing the full range of information from the very general to information at the finest level of granularity.

*Media independence*. The framework must handle all potential media types, including text, audio, video, image, etc. and should, in principle, provide common mechanisms for handling all of them. The framework will rely on existing or developing standards for representing multi-media.

*Semantic adequacy*. Representation structures must have a formal semantics, including definitions of logical operations. There must exist a centralized way of sharing descriptors and information categories

*Incrementality*. The framework must provide support for various stages of input interpretation and output generation, both during annotation (which may be accomplished at different times and with different software) and use. It must also provide for the representation of partial/under-specified results and ambiguities, alternatives, etc. and their merging and comparison.

*Separability*. As a complement to incrementality, it must be possible for NLP applications to easily separate or extract annotation types specific to the task at hand.

*Uniformity*. Representations must utilize same "building blocks" and the same methods for combining them.

*Openness*. The framework must not dictate representations dependent on a single linguistic theory.

*Extensibility*. The framework must provide ways to declare and interchange extensions to the centralized data category registry.

*Human readability*. Representations must be human readable, at least for creation and editing.

*Processability (explicitness)*. Information in an annotation scheme must be explicit—that is, the burden of interpretation should not be left to the processing software.

*Consistency*. Different mechanisms should not be used to indicate the same type of information.

To fulfill these requirements, it is necessary to identify a consistent underlying *data model* for data and its annotations. A data model is a formalized description of the data objects (in terms of composition, attributes, class membership, applicable procedures, etc.) and relations among them, independent of their instantiation in any particular form. A data model capable of capturing the structure and relations in diverse types of data and annotations is a pre-requisite for developing a common corpus-handling environment: it impacts the design of annotation schema, encoding formats and data architectures, and tool architectures.

As a starting assumption, we can conceive of an annotation as a one- or two-way link between an annotation object and a point (or a list/set of points) or span (or a list/set of spans) within a base data set. Links may or may not have a semantics--i.e., a type--associated with them. Points and spans in the base data may themselves be objects, or sets or lists of objects. We make several observations concerning this assumption:

- the model assumes a fundamental linearity of objects in the base,[3] e.g., as a time line (speech); a sequence of characters, words, sentences, etc.; or pixel data representing images;
- the *granularity* of the data representation and encoding is critical: it must be possible to uniquely point to the smallest possible component (e.g., character, phonetic component, pitch signal, morpheme, word, etc.);
- an annotation scheme must be mappable to the structures defined for annotation objects in the model;
- the encoding scheme must be able to capture the object structure and relations expressed in the model, including class membership and inheritance, therefore requiring a sophisticated means to specify linkage within and between documents;
- it is necessary to consider the logistics of identifying spans by enclosing them in start and

---

[3] Note that this observation applies to the *fundamental* structure of stored data. Because the targets of a relation may be either individual objects, or sets or lists of objects, information with more than one dimension is accommodated.

end tags (thus enabling hierarchical grouping of objects in the data itself), vs. explicit addressing of start and end points, which is required for read-only data;

- it must be possible to represent objects and relations in some (fairly straightforward) form that prevents information loss;
- it should be possible to represent the objects and relations in a variety of formats suitable to different tools and applications.

ISO TC37/SC 4's goal is to develop a framework for the design and implementation of linguistic resource formats and processes in order to facilitate the exchange of information between language processing modules. A well-defined representational framework for linguistic information will also provide for the specification and comparison of existing application-specific representations and the definition of new ones, while ensuring a level of interoperability between them. The framework should allow for variation in annotation schemes while at the same time enabling comparison and evaluation, merging of different annotations, and development of common tools for creating and using annotated data. For this purpose we envisage a common "pivot" format based on a data model capable of capturing all types of information in linguistic annotations, into and out of which site-specific representation formats can be transduced. This strategy is similar to that adopted in the design of languages intended to be reusable across platforms, such as Java. The pivot format must support the communication among all modules in the system, and be adequate for representing not only the end result of interpretation, but also intermediate results.

## 4 Terms and definitions

The following terms and definitions are used in the discussion that follows:

**Annotation:** The process of adding linguistic information to language data ("annotation of a corpus") or the linguistic information itself ("an annotation"), independent of its representation. For example, one may annotate a document for syntax using a LISP-like representation, an XML representation, etc.

**Representation**: The format in which the annotation is rendered, e.g. XML, LISP, etc. independent of its content. For example, a phrase structure syntactic annotation and a dependency-based annotation may both be represented using XML, even though the annotation information itself is very different.

**Types of Annotation**: We distinguish two fundamental types of annotation activity:

1. *segmentation* : delimits linguistic elements that appear in the primary data. Including

    - continuous segments (appear contiguously in the primary data)
    - super- and sub-segments, where groups of segments will comprise the parts of a larger segment (e.g., a contiguous word segments typically comprise a sentence segment)
    - discontinuous segments (linking continuous segments)
    - landmarks (e.g time stamps) that note a point in the primary data

    In current practice, segmental information may or may not appear in the document containing the primary data itself. Documents considered to be *read-only,* for example, might be segmented by specifying byte offsets into the primary document where a given segment begins and ends.

2. *linguistic annotation:* provides linguistic and/or communicative information about the segments in the primary data, e.g., a morpho-syntactic annotation in which a part of speech and lemma are associated with each segment in the data. Note that the identification of a segment as a word, sentence, noun phrase, etc. also constitutes linguistic annotation.

    In current practice, when it is possible to do so, segmentation and identification of the linguistic role or properties of that segment are often combined (e.g., syntactic bracketing, or delimiting each word in the document with an XML tag that identifies the segment as a word, sentence, etc.).

**Stand-off annotation:** Annotations layered over a given primary document and instantiated in a document separate from that containing the primary data. Stand-off annotations refer to specific locations in the primary data, by addressing byte offsets, elements, etc. to which the annotation applies. Multiple stand-off annotation documents for a given type of annotation can refer to the same primary document (e.g., two different part of speech annotations for a given text). There is no requirement that a single XML-compliant document may be created by merging stand-off annotation documents with the primary data; that is, two annotation documents may specify trees over the primary data that contain overlapping hierarchies.

## 5  Design principles

The following general principles guide the LAF development:

- The data model and document form are distinct but mappable to one another
- The data model is parsimonious, general, and formally precise
- The data model is built around a clear separation of structure and content
- There is an inventory of logical operations supported by the data model, which define its abstract semantics
- The document form is largely under user control
- The mapping between the flexible document form and data model is via a rigid dump-format
- The mapping from document form to the dump format is documented in an XML Schema (or the functional equivalent thereof) associated with the document
- Mapping is operationalized *either* via schema-based data-binding process *or* via schema-derived stylesheet mapping between the user document and the dump-format document.
- It must be possible to isolate specific layers of annotation from other annotation layers or the primary (base) data; i.e., it must be possible to create a format using stand-off annotation
- The dump format must be designed to enable stream marshalling and unmarshalling

The overall architecture of LAF as dictated by these principles is given in Figure 1. The fundamental principle is that the user controls the representation format for linguistic annotations, which is mappable to the data model. This mapping is accomplished via a rigid "dump" format, isomorphic to the data model and intended primarily for machine rather than human use. The left side of the diagram represents the user-defined document form, and is labeled "human" to indicate that creation and editing, of the resource is accomplished via human interaction with this format. This format should, to the extent possible, be human readable. We will support XML for these formats (e.g., by providing style sheets, examples, etc.) but not disallow other formats. The right side represents the dump format, which is machine processable, and may not be human readable, as it is intended for use only in processing.

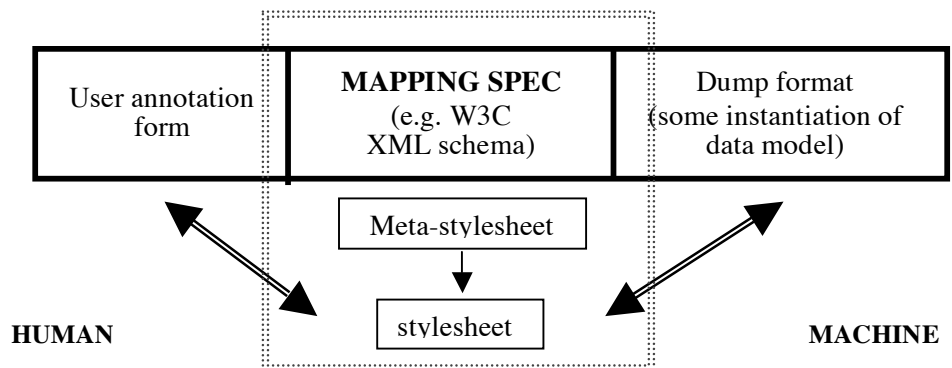

Figure 1. Overall LAF architecture

# 6 Practice

The following set of practices will guide the implementation of the LAF:

- The data model is essentially a feature structure graph with a moderate admixture of algebra (e.g. disjunction, sets), grounded in *n*-dimensional regions of primary data and literals.
- The dump format is isomorphic to the data model.
- Semantic coherence is provided by a registry of features in an XML-compatible format (e.g., RDF), which can be used directly in the user-defined formats and is always used with the dump format.
- Resources will be available to support the design and specification of document forms, for example:
  - XML Schemas in several normal forms based on type definitions and abstract elements that can be exploited via type derivation and/or substitution group;
  - XPointer design-patterns with standoff semantics;
  - Schema annotations specifying mapping between document form and data model;
  - Meta-stylesheet for mapping from annotated XML Schema to mapping stylesheets;
  - Data-binding stylesheets with language-specific bindings (e.g. Java).
- Users may define their own data categories or establish variants of categories in the registry. In such cases, the newly defined data categories will be formalized using the same format as definitions available in the registry, and will be associated with the dump format.
- The responsibility of converting to the dump format is on the producer of the resource.
- The producer is responsible for documenting the mapping from the user format to the data model.
- The ISO working group will provide test suites and examples following these guidelines:
  - The example format should illustrate use of data model/mapping
  - The examples will show both the left (human-readable) and right (machine processable) side formats
  - Examples will be provided that use existing schemes

# 7 Data Model

The data model is built around a clear separation of the *structure* of annotations and their *content*, that is, the linguistic information the annotation provides. The model therefore combines a structural *meta-model*, that is, an abstract structure shared by all documents of a given type (e.g. syntactic annotation), and a set of *data categories* associated with the various components of the structural meta-model.

The structural component of the data model is a directed graph of feature structures capable of referencing *n*-dimensional regions of primary data as well as other annotations. The choice of this model is indicated by its almost universal use in defining general-purpose annotation formats, including the Generic Modeling Tool (GMT) (Ide and Romary, 2001, 2002) and Annotation Graphs (Bird and Liberman, 2001). A small inventory of logical operations over annotation structures is specified, which define the model's abstract semantics. These operations allow for expressing the following relations among annotation fragments:

- *Parallelism:* two or more annotations refer to the same data object;
- *Alternatives:* two or more annotations comprise a set of mutually exclusive alternatives (e.g., two possible part-of-speech assignments, before disambiguation);
- *Aggregation:* two or more annotations comprise a list (ordered) or set (unordered) that should be taken as a unit.

The feature structure graph contains elementary structural nodes to which one or more data category/value pairs are attached, providing the semantics of the annotation.

# 8 LAF Implementation

As specified by the LAF architecture, the dump format will implement a feature structure graph instantiated in XML. All annotations are stand-off--i.e., in documents separate from the primary data and other annotations—in order to support the requirements for *incrementality* and *separability* (see section 3.2).

The XML-based GMT will serve as a starting point for defining the dump format. The GMT implements a simple hierarchical structural model, with mechanisms to support long-distance dependencies, together with a basic set of features structure construction operators for conjunction, disjunction, etc. As such, it is sufficiently expressive to represent the information required in LAF. For examples of GMT application to different annotation types, see Ide, *et al.*, 2000 (terminology, dictionaries and other lexical data); Ide and Romary, 2002 (morphological annotation); and Ide and Romary, 2001b, 2003 (syntactic annotation).

The final implementation of the dump format may differ from the GMT, in particular by mapping its "plain" XML format to RDF/RDFS/OWL.

## 8.1 Data Categories and the Data Category Registry

The central component of the LAF architecture is a Data Category Registry that will contain pre-defined data elements and schemas that can be used directly in annotations. Alternatively, users may define their own data categories or establish variants of categories in the registry; in such cases, the newly defined data categories will be formalized using the same format as definitions available in the registry.

We define a *data category* as an elementary descriptor used in a linguistic annotation scheme. In feature structure terminology, data categories include both attributes (hereafter called *type descriptors*) such as SYNTACTIC CATEGORY and GRAMMATICAL GENDER, as well as a set of associated atomic *values* taken by such attributes, such as NOUN and FEMININE. In both cases we distinguish between the abstraction (concept) behind an attribute or value, and its realization as some string of characters or other object. Figure 1 provides an overview of these relationships. Whereas there is only one concept for a given attribute or value, there may be multiple instantiations.

| type descriptor | value | |
|---|---|---|
| GENDER | MASCULINE FEMININE NEUTER | *conceptual dimension* |
| gen | {m,f,n} | *instantiation* |
| genre | {masc, fem, neut} | *instantiation* |

Figure 1. Data category overview

The DCR under development within ISO TC37 SC4 is built around this fundamental concept/instance distinction. In principle, the DCR provides a set of reference concepts, while the annotator provides a *Data Category Specification* (DCS) that comprises a mapping between his or her scheme-specific instantiations and the concepts in the DCR. As such, the DCS provides documentation for the linguistic annotation scheme in question. The DCS for a given annotation document/s is included or referenced in any data exchange to provide the receiver with the information required to interpret the annotation content or to map it to another instantiation. Semantic integrity is guaranteed by mutual reference to DCR concepts.

The DCR is intended to provide a set of formally-defined reference categories. "Formal definition" in this context includes natural language definitions for each category accompanied by specification of the possible values each category may take. At present, we envision instantiation of the DCR as a simple database in which each entry is either a type descriptor or value. Data categories will be referenced either by the DCR entry identifier, or, since the DCR will be publicly available on-line, via a URI.

Note that this simple instantiation of the DCR makes no distinction in terms of representation between type descriptors and values; each is considered as a data category and provided with an entry identifier for reference. Only minimal constraints on their use in an annotation are specified--i.e., constraints on descriptor/value combinations given in the descriptor entry. The broader structural integrity of an annotation is provided by placing constraints on nodes in the annotation graph (as defined in the LAF architecture) with which a given category can be associated. For example, the structural graph for a syntactic constituency analysis would consist of a hierarchy of typed nodes corresponding to the non-terminals in the grammar, with constraints on their embedding, and with which only appropriate descriptor/value pairs may be associated. Node types

(e.g., NP, VP) as well as associated grammatical information (e.g., tense, number) may all be specified with data categories drawn from the DCR.

A more formal specification of data categories could be provided using mechanisms such as RDF Schema (RDFS) and the Ontology Web Language (OWL) to formalize the properties and relations associated with data categories. For example, consider the following RDF Schema fragment:

```
<rdfs:Class rdf:about="#Noun">
  <rdfs:label>Noun</rdfs:label>
  <rdfs:comment>Class for
        nouns</rdfs:comment>
</rdfs:Class>
<rdfs:Property rdf:about="#number">
  <rdfs:domain
      rdfs:resource="Noun"/>
  <rdfs:range
      rdf:resource="rdfs:#Literal"/>
</rdfs:Property>
```

This fragment defines a class of objects called " "Noun" that has a property "number". Note that the schema defines the classes but does not instantiate objects belonging to the class; instantiation may be accomplished directly in the annotation file, as follows (for brevity, the following examples assume appropriate namespace declarations specifying the URIs of schema and instance declarations):

```
<Noun rdf:about="Mydoc#W1">
    <number rdf:value="Plural"/>
</Noun>
```

where "Mydoc#W1" is the URI of the word being annotated as a noun. Alternatively, the DCR could contain instantiations of basic data elements, specifying values for properties, which can be referenced directly in the annotation:

```
<Noun rdf:ID="NMP">
    <number rdf:value="plural"/>
</Noun>
```

The annotation file could then reference the pre-defined instance as follows:

```
<rdf:Description rdf:about="myDoc#W1">
    <POS rdf:resource="categories#NMS"/>
</rdf:Description>
```
[4]

The class and sub-class mechanisms provided in RDFS and its extensions in OWL can also be used to model the structure of annotations—that is, to identify and constrain nodes in the annotation graph itself. For example, the hierarchical structure defined by ISLE/MILE for lexical entries (Calzolari, *et al*. 2003) has been modeled in an RDF Schema (Ide, *et al.*, 2003). Another example is the time ontology[5] developed by, the DAML[6] effort, which reflects the internal structure of time and event annotations.

An RDFS/OWL specification of data categories would enable greater control over descriptor/value use and also allow for the possibility of inferencing over annotations. However, it would also demand definition of a precise hierarchy of linguistic categories and a distinction between classes (objects) and properties that could place unwanted constraints on annotation form and content. Therefore, any such specification of data categories is left to the annotator, at least for the time being.

It is anticipated that many annotators will use their own category names and values and provide a mapping to DCR categories. The DCR will include an XML template for specifying this mapping, as well as for defining variants and new descriptor/value pairs.

The Data Category Registry will support multiple languages by providing the following:

- reference definitions for data categories in various languages;

- data element names for the data categories in various languages;

- description of usage in language-specific contexts, including definitions, usage notes, examples, and/or lists of values (e.g., GENDER takes the values *masculine*, *feminine* in French; *masculine*, *feminine*, *neuter* in German)

---

[4] In these examples, NUMBER is given literal values. However, with OWL it is possible to restrict the range of possible values by enumeration.

[5] Accessible from http://www.cs.rochester.edu/~ferguson/daml/.
[6] http://www.daml.org

The goal of the Data Category Registry is not to impose a specific set of categories, but rather to ensure that the semantics of data categories included in annotations (whether they exist in the Registry or not) are well-defined and understood. It is possible that several different instantiations of the same category (e.g., noun) and/or different schemas describing the same phenomenon could exist in the registry, to be used as desired by annotators. The purpose of the registry is solely to gather together (and where necessary, harmonize) existing schemas and instances in use by the language technology community as a resource for the annotation of linguistic data. The formally defined set of categories will have several functions: (1) it will provide a precise semantics for annotation categories that can be either used "off the shelf" by annotators or modified to serve specific needs; (2) it will provide a set of reference categories onto which scheme-specific names can be mapped; and (3) it will provide a point of departure for definition of variant or more precise categories.

## 9 Conclusion

In this paper we describe the Linguistic Annotation Framework under development by ISO TC37/SC 4 WG1. Its design is intended to allow for, on the one hand, maximum flexibility for annotators, and. on the other, processing efficiency and reusability. This is accomplished by separating user annotation formats from the exchange/processing format. This separation ensures that pre-existing annotations are compatible with LAF, and that users have the freedom to design specific schemes to meet their needs, while still conforming to LAF requirements.

LAF provides for the use of any annotation format consistent with the feature structure-based data model that will be used to define the dump format. This suggests a future scenario in which annotators may create and edit annotations in a proprietary format, transduce the annotations using available tools to the dump format for interchange and/or processing, and if desired, transduce the dump form of the annotations (and/or additional annotation introduced by processing) back into the proprietary format. We anticipate the future development of annotation tools that provide a user-oriented interface for specifying annotation information, and which then generate annotations in the pivot format directly. Thus the pivot format is intended to function in the same way as, for example, Java byte code functions for programmers, as a universal "machine language" that is interpreted by processing software into an internal representation suited to its particular requirements. As with Java byte code, users need never see or manipulate the pivot format; it is solely for machine consumption.

Part of the work of SC4 WG1 is to provide development resources, including schemas, design patterns, and stylesheets, which will enable annotators and software developers to immediately adapt to LAF. Example mappings, e.g., for XCES-encoded annotations, will also be provided. In this way, we hope to realize the goal of harmonized and reusable resources in the near future.

For general information on the work of the ISO committee on language resources, consult the ISO TC37/SC4 website (http://www.tc37sc4.org).